\documentclass[runningheads]{llncs}

\usepackage{accv}

\usepackage{accvabbrv}

\usepackage{graphicx}
\usepackage{booktabs}
\usepackage{bm}
\usepackage[accsupp]{axessibility}  

\usepackage[pagebackref,breaklinks,colorlinks,citecolor=accvblue]{hyperref}

\usepackage{orcidlink}
\usepackage{makecell, booktabs, multirow, colortbl}
\usepackage{color}

\begin{document}

\title{AnolisVLA: \\ Lightweight Vision-Language-Action Model \\ with Deep State Space Models \\for Mobile Manipulation} 

\titlerunning{AnolisVLA: Lightweight VLA with Deep SSM}

\author{Yusuke Takagi \orcidlink{0009-0004-6687-7929} \and
Motonari Kambara \orcidlink{0000-0002-1991-9119} \and
Daichi Yashima\orcidlink{0009-0005-2087-2038} \and \\ Koki Seno  \orcidlink{0009-0003-6971-8512} \and  Kento Tokura \orcidlink{0009-0009-4524-9022}  \and Komei Sugiura\orcidlink{0000-0002-0261-0510}}

\authorrunning{Y.~Takagi et al.}

\institute{Keio University, Japan \\
\email{
\{yusuke.10.06, motonari.k714, ydaichi1207, \\koki.seno, tkento1985, komei.sugiura\}@keio.jp
}
}

\maketitle


\begin{abstract}
In this study, we address the problem of language-guided robotic manipulation, where a robot is required to manipulate a wide range of objects based on visual observations and natural language instructions. 
This task is essential for service robots that operate in human environments, and requires safety, efficiency, and task-level generality. 
Although Vision-Language-Action models (VLAs) have demonstrated strong performance for this task, their deployment in resource-constrained environments remains challenging because of the computational cost of standard transformer backbones. 
To overcome this limitation, we propose \textit{AnolisVLA }, a lightweight VLA that uses a deep state space model to process multimodal sequences efficiently.
The model uses lightweight, fast sequential state modeling to process visual and textual inputs. This design enables efficient trajectory generation.
We evaluated the proposed method in simulation and on two real robot platforms.
Compared with SmolVLA at a comparable parameter scale, AnolisVLA achieved higher task success rates, faster inference, and comparable GPU memory usage.
In mobile manipulation, it outperformed the larger, 3-billion-parameter $\pi_{0.5}$ model by 21 percentage points in task success rate while achieving approximately three times faster inference.

\end{abstract}
\begin{figure}[tb]
  \centering
  \includegraphics[width=\textwidth,height=4.5cm,keepaspectratio]{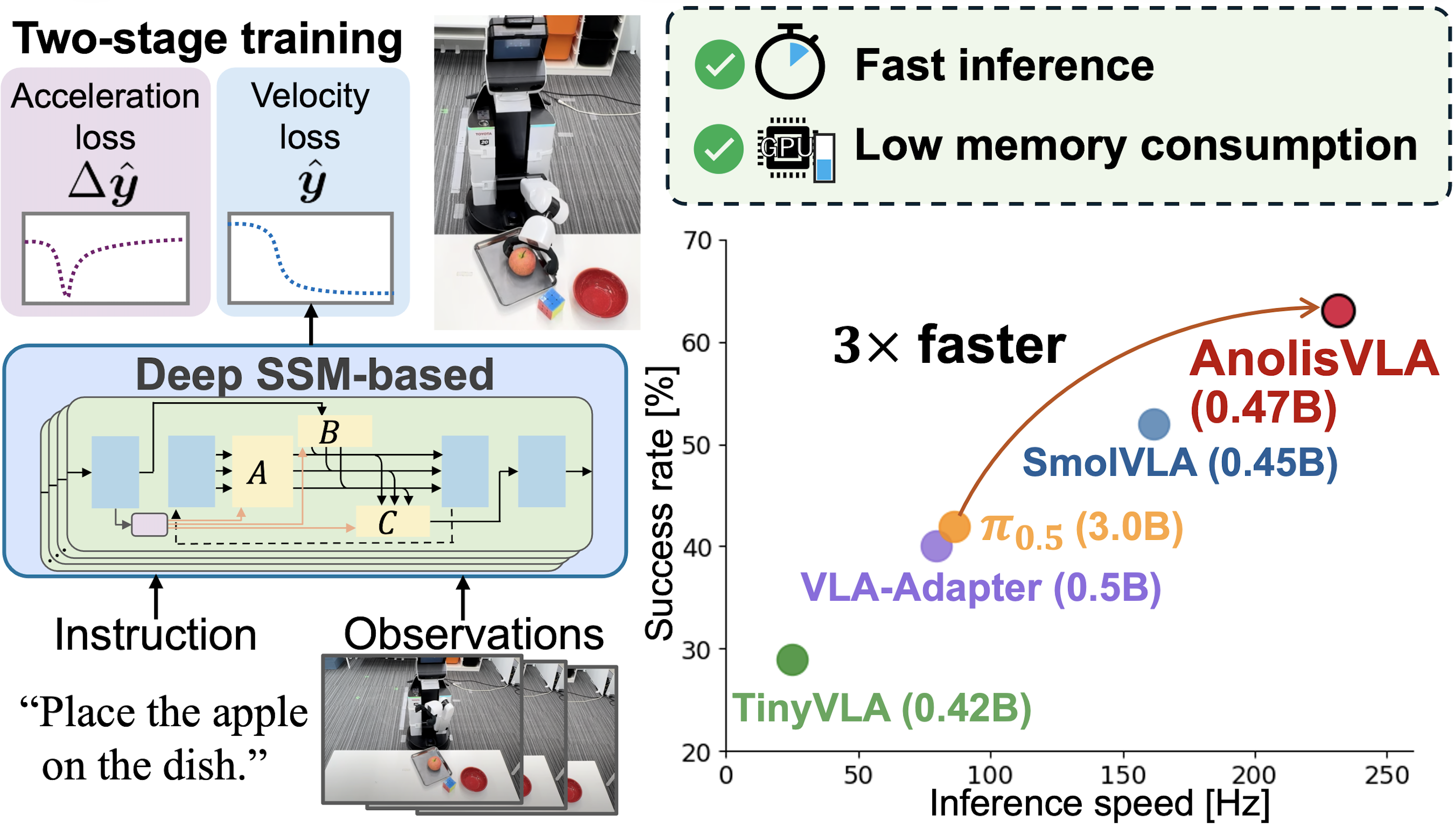}
  \caption{Overview of AnolisVLA and real-world performance. (Left) Our deep SSM backbone processes language instructions and robot observations to generate trajectories, leveraging a two-stage training strategy with the acceleration loss to improve temporal consistency in action sequences. (Right) Real-world mobile manipulation experiment results. The horizontal and vertical axes represent inference speed and the average success rate, respectively.}
  \label{fig:eye-catch}
\end{figure}
\section{Introduction}
\label{sec:intro}
Object manipulation is essential for service robots that operate in everyday environments, and requires safety, efficiency, and task-level generality.
In such environments, robots are often instructed by users through natural language, which makes language-conditioned manipulation a practical necessity.
Based on this background, Vision-Language-Action models (VLAs) have been studied extensively.
Despite significant advances in VLAs \cite{pi05, UniVLA, Kim2024openvla}, deploying them on real-world robots remains challenging because their backbones require substantial memory and compute at inference time.

Given the above background, in this study, we address language-guided robotic manipulation.
%
A typical use case involves a robot performing pick-and-place tasks in a cluttered tabletop environment.
For instance, as shown in Fig.~\ref{fig:eye-catch}, given the instruction ``Place the apple in the dish,'' the robot grounds the referring expressions in the instruction to the specific objects in the scene to identify the target apple among various distractors and moves it to the designated container.

Although VLAs~\cite{Brohan2023rt2, Kim2024openvla, pi0, pi05, UniVLA, X-VLA} demonstrate strong performance in such manipulation tasks, 
deploying them under resource constraints remains challenging, because their backbone constitutes a major source of inference latency and memory consumption.
For example, OpenVLA builds on Llama-2 7B~\cite{Kim2024openvla}, and even the comparatively compact $\pi_{0.5}$ uses a 3B-parameter language model~\cite{pi05}.
%
A key bottleneck is backbone sequence modeling based on transformers.
Transformer self-attention scales quadratically with sequence length, which becomes problematic because VLAs typically concatenate vision, language, and robot-state tokens into a single sequence.
As the number of vision tokens or temporal context grows, this quadratic cost can be a critical bottleneck for real-time deployment on resource-constrained platforms.

To overcome this limitation, we propose \textit{AnolisVLA }, a lightweight VLA that leverages a deep state space model (deep SSM) for efficient multimodal sequence modeling.
By exploiting the linear complexity of deep SSMs with respect to sequence length, AnolisVLA mitigates the inference bottlenecks.
{
%
Concretely, AnolisVLA replaces the transformer backbone with a deep SSM backbone that processes proprioceptive, visual, and language tokens through recurrent state updates.
This design preserves the multimodal tokenization and action decoding pipeline while reducing the cost of sequence modeling.
}
Additionally, unlike many existing VLAs~\cite{Brohan2023rt2, Kim2024openvla} that discretize continuous actions into tokens to use the language model's autoregressive decoding, AnolisVLA directly generates continuous actions from the final hidden state of the backbone.

The main contributions of this study are summarized as follows:
\begin{enumerate}
    \item We propose AnolisVLA, a lightweight vision-language-action model that reduces inference latency and VRAM usage for resource-constrained robotic deployment.
    \item We demonstrate that deep SSMs enable efficient VLA modeling while providing a better performance-efficiency tradeoff than transformer-based VLAs of comparable scale.
    \item We introduce a two-stage trajectory learning strategy and acceleration supervision to capture both action magnitudes and step-to-step temporal changes in continuous action sequences.
\end{enumerate}

\section{Related Work}
\subsection{Vision-Language-Action Models}
Pre-trained vision-language models have been adopted increasingly in robotics~\cite{pi05, VoxPoser, VLMPC, Hori}.
As surveyed in~\cite{Ma2024, Kawaharazuka2023}, VLAs have emerged as one of the most prominent approaches for language-conditioned manipulation~\cite{pi0, pi05, VLA-Adapter, TinyVLA, SmolVLA}, where a pre-trained vision-language model is fine-tuned on robot demonstrations~\cite{Brohan2023rt2, Kim2024openvla, octo, pi05}.
However, these models are built on multimodal large language models (MLLMs), often exceeding 7B parameters~\cite{Brohan2023rt2, Kim2024openvla}, and are slow during inference, which makes real-world deployment difficult~\cite{TinyVLA, SmolVLA}.
To address these issues, several works investigated smaller VLAs that reduce inference cost~\cite{VLA-Adapter, TinyVLA, SmolVLA}.

In addition to architectural design, the choice of training data is a key factor in VLA performance.
Open X-Embodiment (OXE)~\cite{OXE} aggregates trajectory data from multiple robot platforms, including Fractal~\cite{RT-1} and BridgeData V2~\cite{BridgeV2}, and includes many manipulation episodes paired with language instructions.
OXE has become a standard training source for many VLAs~\cite{OXE, Kim2024openvla, SpatialVLA, CoT-VLA, TraceVLA}.
For evaluation, simulation benchmarks are widely used because they offer reproducibility at lower cost.
Meta-World~\cite{Yu2019MetaWorld} defines diverse manipulation tasks on a single robot arm and has been adopted in several studies~\cite{TinyVLA, SmolVLA}.

\subsection{Deep State Space Models}

Deep SSMs model sequences via a recurrent state that is updated at each timestep, which offers linear complexity in sequence length as an alternative to the quadratic cost of transformer self-attention~\cite{gu2021efficiently, gu2022parameterization}.
S4 introduced structured parameterizations to handle long-range dependencies \cite{gu2021efficiently} and Mamba further introduced input-dependent selection mechanisms that control information propagation over the sequence \cite{gu2024mamba}.
These backbones have been adopted in multimodal language models such as Cobra, VL-Mamba, and EMMA, with an emphasis on efficient long-context processing\cite{Zhao2025cobra, Qiao2024vlmamba, Xing2025emma}.

In robotics, deep SSMs have been adopted in various ways depending on where state space recurrence is placed in the learning and control pipeline.
Some VLAs use deep SSM backbones to map multimodal tokens to end-effector representation.
For example, RoboMamba predicts contact points and end-effector pose with a lightweight head \cite{Robomamba}.
Deep SSMs have also been used as multimodal fusion modules. FlowRAM integrates Mamba blocks within a conditional flow matching pipeline \cite{flowram}.
Other studies (e.g. \cite{lumos,robossm}) used state space recurrence for world models or for long-horizon imitation learning.

Another research direction involves the study of deep SSM backbones for visuo-motor policy learning, where policies map sensory inputs to end-effector trajectory representation under imitation learning or generative objectives.
This includes Mamba-based diffusion policies and proprioceptive motion encoders \cite{MaIL,cao2025mamba,oh2024dispo,tsuji2025mamba}.

Compared with these visuo-motor and generative policies, AnolisVLA differs in two key aspects.
First, we train for temporal consistency in action sequences by supervising both the end-effector trajectory and temporal differences with a two-stage training strategy.
Second, AnolisVLA generates a short-horizon continuous action chunk \cite{ACT} from multimodal tokens in a single forward pass, without action discretization or iterative inference.
As a result, AnolisVLA targets language-guided manipulation settings where continuous trajectory generation and efficient inference are both important.
\section{Problem Statement}
In this study, we focus on language-guided robotic manipulation.
The input to the task consists of images, proprioceptive state, and a natural language instruction.
The output is the robot's trajectory, formulated as a sequence of actions $\hat{\bm{y}}^{(1:H)} \in \mathbb{R}^{d_a \times H}$, where $d_a$ and $H$ denote the dimension of the action space and the action chunking length, respectively.
We use the success rate as the main evaluation metric, and additionally report inference time and inference-time VRAM usage in the real-world experiments.

\section{Methodology}
\begin{figure}[tb]
  \centering
  \includegraphics[width=\textwidth,height=6.5cm,keepaspectratio]{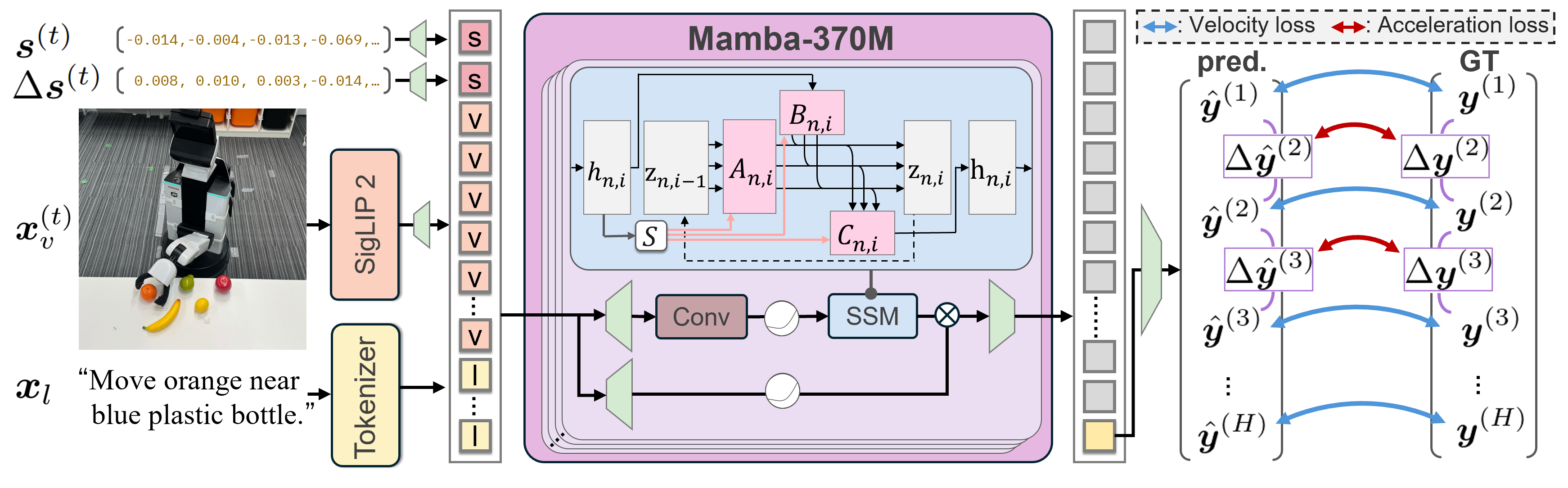}
  \caption{
    Model architecture of AnolisVLA. Multimodal tokens, including proprioception, state delta, vision, and language, are concatenated and processed by a Mamba backbone, and the final token predicts an $H$-step action chunk. 
    The two-stage training supervises action commands and their temporal differences.
    Here, $\bm{s}^{(t)}$, $\Delta \bm{s}^{(t)}$, $\bm{x}^{(t)}_v$, and $\bm{x}_l$ denote the state, state delta, visual observation, and language instruction at time step $t$, respectively.
    On the right-hand side, $\hat{\bm{y}}$ and $\bm{y}$ denote the predicted future actions and ground truth, while $\Delta \hat{\bm{y}}$ and $\Delta \bm{y}$ denote their temporal differences.
}
  \label{fig:model}
\end{figure}
AnolisVLA is closely related to Mamba~\cite{gu2024mamba}-based MLLMs, including Cobra \cite{Zhao2025cobra}, VL-Mamba \cite{Qiao2024vlmamba}, and EMMA \cite{Xing2025emma}.
The architectural frameworks of these models can be adapted to many existing VLAs \cite{pi0, pi05, VLA-Adapter}.
Because both Mamba and transformers operate as causal sequence models over a shared token space, Mamba can seamlessly replace the transformer backbone while preserving the upstream multimodal tokenization and downstream action decoding pipelines.
\subsection{Preliminaries: Mamba}

Deep SSMs have become effective substitutes for transformers because of their linear scalability with sequence length $L$.
In particular, transformer self-attention scales quadratically as $\mathcal{O}(L^2)$, whereas deep SSM-based models process sequences in $\mathcal{O}(L)$ time, which enables efficient long-context modeling.

Mamba~\cite{gu2024mamba} is a selective deep SSM that introduces an input-dependent selection mechanism, which allows the model to dynamically propagate or forget information along the token sequence.
We adopt Mamba as the backbone in AnolisVLA 's architecture while keeping the same multimodal tokenization and action decoding pipeline.

\subsection{Architecture}
Fig.~\ref{fig:model} illustrates the architecture of AnolisVLA , which consists of three components: an input embedding module, a Mamba backbone, and an action head.

\noindent{\textbf{Input embedding module.}}
We define the input $\bm{x}^{(t)}$ at time step $t$ as follows:
\begin{equation}
    \bm{x}^{(t)} = \{ \bm{s}^{(t)}, \Delta\bm{s}^{(t)}, \bm{x}^{(t)}_v, \bm{x}_l \},
\end{equation}
where $\bm{s}^{(t)}$, $\Delta\bm{s}^{(t)}(:= \bm{s}^{(t)} - \bm{s}^{(t-1)})$, $\bm{x}^{(t)}_v$, and $\bm{x}_l$ represent the robot state, delta state, RGB images, and natural language instruction, respectively.
The input embedding module projects each component of $\bm{x}^{(t)}$ into a shared latent space of dimension $D$.
Specifically, $\bm{s}^{(t)}$ and $\Delta\bm{s}^{(t)}$ are each mapped via linear projections to single-token embeddings $\bm{h}^{(t)}_s \in \mathbb{R}^{D}$ and $\bm{h}^{(t)}_{\Delta s} \in \mathbb{R}^{D}$, respectively.
$\bm{x}^{(t)}_v$ is encoded using SigLIP 2~\cite{siglip_2}, initialized with pre-trained weights, and fine-tuned end-to-end on robot demonstration data, followed by a linear projection, which yields visual tokens $\bm{h}^{(t)}_v \in \mathbb{R}^{N_v \times D}$, where $N_v$ denotes the visual tokens' sequence length.
Then $\bm{x}_l$ is tokenized and projected by a linear layer to obtain $\bm{h}_l \in \mathbb{R}^{N_l \times D}$, where $N_l$ denotes the sequence length.
These embeddings are concatenated to form the input sequence:
\begin{equation}
\bm{H}_0 = [\bm{h}^{(t)}_s,\; \bm{h}^{(t)}_{\Delta s},\; \bm{h}^{(t)}_v,\; \bm{h}_l] \in \mathbb{R}^{L \times D},
\end{equation}
where $L = 2 + N_v + N_l$ is the total number of tokens, with the constant $2$ accounting for $\bm{h}^{(t)}_s$ and $\bm{h}^{(t)}_{\Delta s}$.

\noindent{\textbf{Mamba backbone.}}
The backbone consists of $K$ stacked Mamba blocks.
Each block comprises a selective deep SSM layer, a SiLU-gated linear unit, layer normalization, and a residual connection, following~\cite{gu2024mamba}.
For notational convenience, we index the input sequence positionally as $\bm{H}_0 = [\bm{h}_{0,1}, \dots, \bm{h}_{0,L}]$.
The $n$-th block takes the output of the preceding block, $\bm{H}_{n-1} = [\bm{h}_{n-1,1}, \dots, \bm{h}_{n-1,L}]$, and applies the deep selective SSM update for each token position $i$:
\begin{equation}
\begin{aligned}
\bm{z}_{n,i} &= {\bm{A}}_{n,i}\, \bm{z}_{n,i-1} + {\bm{B}}_{n,i}\, \bm{h}_{n-1,i}, \\
\tilde{\bm{h}}_{n,i} &= \bm{C}_{n,i}\, \bm{z}_{n,i},
\end{aligned}
\end{equation}
where $\bm{z}_{n,i}$ denotes the latent state at the $n$-th block and the $i$-th token position. 
Moreover, ${\bm{A}}_{n,i}$, ${\bm{B}}_{n,i}$, and $\bm{C}_{n,i}$ are system matrices.
The intermediate output $\tilde{\bm{h}}_{n,i}$ is then passed through the SiLU-gated linear unit, layer normalization, and a residual connection to obtain the block output $\bm{h}_{n,i}$.

The recurrent state update can be interpreted as a filtering process that continuously refines a running estimate of task-relevant information as new tokens arrive, aligning naturally with robotic manipulation where the model is required to incrementally integrate visual observations to determine actions. 
Because Mamba processes the sequence causally, token $i$ can only aggregate information from the prefix $\{\bm{h}_{n-1,j}\}_{j\le i}$ through the recurrent hidden state.
This causal structure means that the interaction between modalities is governed by the token ordering defined above.



\noindent{\textbf{Action head.}}
We obtain the predicted action trajectory from the final token embedding of the last block:
\begin{equation}
\hat{\bm{y}}^{(1:H)} = f_\mathrm{head}(\bm{h}_{K,L}),
\end{equation}
where $f_\mathrm{head}(\cdot)$ denotes a linear projection. 
We apply a linear projection to the hidden state at the final token position, as Mamba recurrently aggregates information from the entire input sequence into this hidden state.

\subsection{Loss Function}
Following standard practices~\cite{Kim2024openvla, openvla-oft}, each action $\bm{y}^{(\tau)}$ is modeled as the delta end-effector pose (i.e., the end-effector velocity) at horizon step $\tau$.
Therefore, the temporal difference between consecutive actions, $\Delta \bm{y}^{(\tau)} := \bm{y}^{(\tau)} - \bm{y}^{(\tau-1)}$, can be interpreted as end-effector acceleration.
In trajectory generation, consistency in both velocity and acceleration is important for modeling step-to-step temporal changes in real-world execution.
Motivated by this observation, we design a two-stage loss that supervises both quantities.
In the first stage, we minimize an L1 loss on the predicted velocity (i.e., action) sequence:
\begin{equation}
    \mathcal{L}_{\text{vel}} = \frac{1}{H} \sum_{\tau=1}^{H} \left\lVert \bm{y}^{(\tau)} - \hat{\bm{y}}^{(\tau)} \right\rVert_1.
\end{equation}
In the second stage, we update the objective with an acceleration loss that supervises the temporal differences of the action sequence:

\begin{equation}
\begin{aligned}
    \mathcal{L}_{\text{acc}} &= \frac{1}{H-1} \sum_{\tau=2}^{H} \left\lVert \Delta \bm{y}^{(\tau)} - \Delta \hat{\bm{y}}^{(\tau)} \right\rVert_1,
\end{aligned}
\end{equation}

where $\Delta \hat{\bm{y}}^{(\tau)}$ is the temporal difference of $ \hat{\bm{y}}^{(\tau)}$.
The overall training objective in the second stage is $\mathcal{L}_{\text{vel}} + \lambda_{\text{acc}} \mathcal{L}_{\text{acc}}$, where $\lambda_{\text{acc}}$ is the weight.
This staged training allows the model to first learn an appropriate reconstruction of the reference trajectory before the acceleration term refines its temporal consistency.

\section{Experiments}
\subsection{Settings}

\begin{table}[t] \centering \setlength{\tabcolsep}{5pt} \caption{Quantitative comparison between AnolisVLA and the baseline methods on the Meta-World benchmark. The best results are highlighted in \textbf{bold}, and the second-best results are \underline{underlined}.} 
\label{tab:metaworld_results} \small \begin{tabular}{lc ccccc} \toprule Method & Total Params & Easy & Medium & Hard & Very Hard & Avg. \\ \midrule $\pi_{0.5}$ \cite{pi05} & 3.0\text{B} & 68.20 & 37.30 & 41.70 & \underline{28.00} & 43.80 \\ \midrule VLA-Adapter \cite{VLA-Adapter} & 0.5\text{B} & 3.75 & 0.00 & 0.00 & 0.00 & 0.94 \\ TinyVLA \cite{TinyVLA} & 0.42\text{B} & 77.60 & 21.50 & 11.40 & 15.80 & 31.58 \\ SmolVLA \cite{SmolVLA} & 0.45\text{B} & \underline{82.50} & \underline{41.80} & \underline{45.00} & \textbf{60.00} & \underline{57.33} \\ \rowcolor{green!10} \textbf{AnolisVLA} & 0.47\text{B} & \textbf{92.86} & \textbf{72.73} & \textbf{91.67} & \textbf{60.00} & \textbf{79.32} \\ \bottomrule \end{tabular} 
\end{table}
As a simulation benchmark, we used Meta-World~\cite{Yu2019MetaWorld}, which is widely recognized as a standard suite for language-guided robotic manipulation.
The benchmark comprises a diverse suite of manipulation tasks, including object relocation and door/drawer opening, performed by a simulated Sawyer robot arm in a tabletop setting.
Each sample includes egocentric RGB images, robot state, a natural language instruction, and the reference trajectory.
On this benchmark, $d_a = 4$, which corresponds to the three-dimensional end-effector position and one-dimensional continuous gripper control.

The Meta-World dataset contains 2,500 episodes across 50 tasks, recorded at 80 Hz with an average length of 81.9 frames.
The Easy, Medium, Hard, and Very Hard suites correspond to single-step, multi-step, contact-intensive, and highest-complexity tasks, respectively.
We use 2,000 episodes for training and 500 for validation and hyperparameter selection, and evaluate on the benchmark's predefined evaluation episodes.


\subsection{Quantitative Results}
Table \ref{tab:metaworld_results} shows the quantitative comparison results between AnolisVLA and the baseline methods on the Meta-World benchmark. 
We adopted SmolVLA \cite{SmolVLA}, VLA-Adapter \cite{VLA-Adapter}, and TinyVLA \cite{TinyVLA} as the baseline methods. 
Each uses a transformer-based language model as its backbone at a parameter scale comparable with AnolisVLA.
This setup provides a comparison with transformer-based VLAs at a similar parameter scale.
Given that $\pi_{0.5}$ \cite{pi05} is widely recognized as a standard large-scale VLA, we included it as a baseline method.
We used the success rate as the evaluation metric.

AnolisVLA achieved success rates of 92.86\%, 72.73\%, 91.67\%, and 60.00\% on Easy, Medium, Hard, and Very Hard task suites, respectively. 
Furthermore, AnolisVLA achieved an average success rate of 79.32\% across all task suites.
By contrast, the average success rates of the baseline methods $\pi_{0.5}$, VLA-Adapter, TinyVLA, and SmolVLA were 43.80\%, 0.94\%, 31.58\%, and 57.33\%, respectively.
Notably, AnolisVLA outperformed SmolVLA, the strongest baseline, by 21.99 points, and surpassed $\pi_{0.5}$ by 35.52 points.
These results show that AnolisVLA outperformed the baseline approaches while maintaining a competitive or smaller model size.
\subsection{Qualitative Results}
\begin{figure}[tb]
  \centering
  \includegraphics[  width=\textwidth,
  height=7cm,
  keepaspectratio]{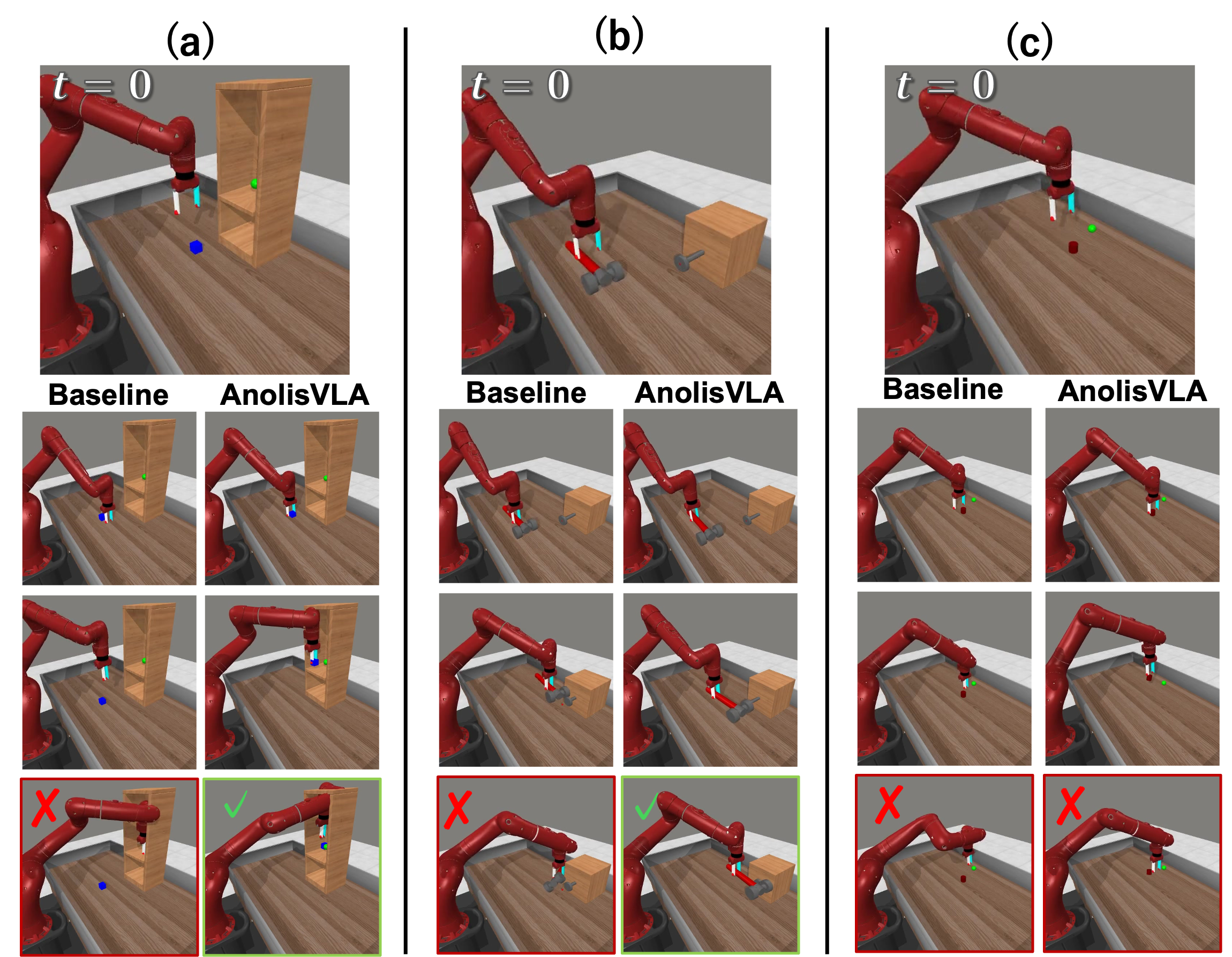}
  \caption{\textbf{Qualitative results.}
        We display the initial observation $\bm{x}_v^{(0)}$ and the chronological sequence of observations $\bm{x}_v^{(t)}$ during task execution.
        In each example, the robot executes the task based on the following instructions: (a) ``Pick and place a puck onto a shelf.'' (b) ``Hammer a screw on the wall.'' (c) ``Push the puck to a goal.''
}
  \label{fig:qualitative}
\end{figure}
Fig. \ref{fig:qualitative} presents the qualitative results of AnolisVLA compared with SmolVLA, which was the strongest baseline. 
Fig. \ref{fig:qualitative}(a) and Fig. \ref{fig:qualitative}(b) illustrate successful episodes of AnolisVLA.
In the episode shown in (a), a blue puck and a wooden shelf were present at $t=0$, and $\bm{x}_l$ was ``Pick and place a puck onto a shelf.''
AnolisVLA successfully grasped the puck and placed it onto the shelf. 
By contrast, although the baseline method approached the vicinity of the puck, it failed to grasp it and moved toward the shelf, which resulted in failure.
In the episode depicted in Fig. \ref{fig:qualitative}(b), a hammer and a screw on a wooden block were present at $t=0$, and $\bm{x}_l$ was ``Hammer a screw on the wall.'' 
AnolisVLA appropriately grasped the hammer and struck the screw. 
Conversely, although the baseline method grasped the hammer, it failed to complete the task because it swung the tool along an incorrect trajectory, thereby missing the screw. 
These results indicate that AnolisVLA was capable of generating complex trajectories that used a tool appropriately to interact with a target object conditioned on language instructions.

Fig. \ref{fig:qualitative}(c) illustrates a failure case of AnolisVLA. 
In this episode, initialized at $t=0$, a red puck and green goal point were present on the table, and $\bm{x}_l$ was ``Push the puck to a goal.''
AnolisVLA failed because it could not make appropriate contact with the puck; it moved the arm past it instead of pushing it toward the goal.
Similarly, the baseline method also failed by missing the target puck entirely.
These results suggest that appropriately localizing the target object and generating precise contact trajectories for achieving tasks remain challenging.

\subsection{Ablation Studies}
\begin{table}[t]
  \centering
  \setlength{\tabcolsep}{7pt}
  \caption{Quantitative results of the ablation studies. The table presents the average success rates on the Meta-World benchmark. The best score is in bold.}
  \label{tab:ablation_sim}
  \begin{tabular}{cccc}
    \toprule
    Model & Backbone & Acceleration loss & Success rate [\%] \\
    \midrule
    (i)  & SmolLM (Transformer) & \checkmark & 76.00 \\
    (ii)  & Mamba-370M (Mamba) &  & 74.59 \\
    \rowcolor{green!10}
    (iii) & Mamba-370M (Mamba) & \checkmark & \textbf{79.32} \\
    \bottomrule
  \end{tabular}
\end{table}
To validate the effectiveness of AnolisVLA , we conducted ablation studies on the backbone and the acceleration loss. Table~\ref{tab:ablation_sim} shows the results on Meta-World. The table reports the average success rate across the task suites.

\noindent{\textbf{{Backbone Architecture Ablation.}}

To investigate whether the proposed Mamba backbone provides benefits beyond model scaling, we compared it with a lightweight transformer backbone under a comparable parameter scale.
Specifically, we evaluated two variants that share the same multimodal tokenization, action head, and acceleration-loss training scheme, but differ only in the backbone.
We selected SmolLM as the transformer backbone because it is closely aligned with the lightweight language-model backbone used in SmolVLA, which achieved the strongest performance among the small-scale transformer-based baselines in our experiments.
The success rates of model (iii), which uses the Mamba-370M backbone, and model (i), which uses the transformer backbone, were 79.32\% and 76.00\%, respectively.
This shows that the Mamba backbone improves the average success rate by 3.32 points while maintaining a comparable parameter scale.
An additional evaluation over five random seeds showed that the Mamba backbone achieved a significantly higher success rate than the transformer backbone ($p < 0.05$).

\noindent{\textbf{Acceleration Loss Ablation.}}
We further evaluated the effect of the acceleration loss by comparing model (ii) and model (iii), both of which use the same Mamba-370M backbone. 
The success rates of model (ii), trained without the acceleration loss, and model (iii), trained with the acceleration loss, were 74.59\% and 79.32\%, respectively.
This result indicates that supervising temporal differences in the action sequence improved the average success rate by 4.73 points. 
These results suggest the contributions of both the Mamba backbone and the acceleration loss.

\section{
    Real-World Experiments
    \label{sec:real-world}
}
We validated AnolisVLA in real-world experiments using two real robot platforms shown in Fig. \ref{fig:setting}: a mobile manipulator and a tabletop bimanual manipulator.
We evaluated whether the model can follow natural language instructions and generate trajectories in real-world settings, and reported task success rates, inference time per action chunk, and inference-time VRAM usage under limited compute and memory budgets.
\subsection{Settings}
\begin{figure}[tb]
  \centering
  \includegraphics[  width=0.95\textwidth,
  height=3.6cm,
  keepaspectratio]{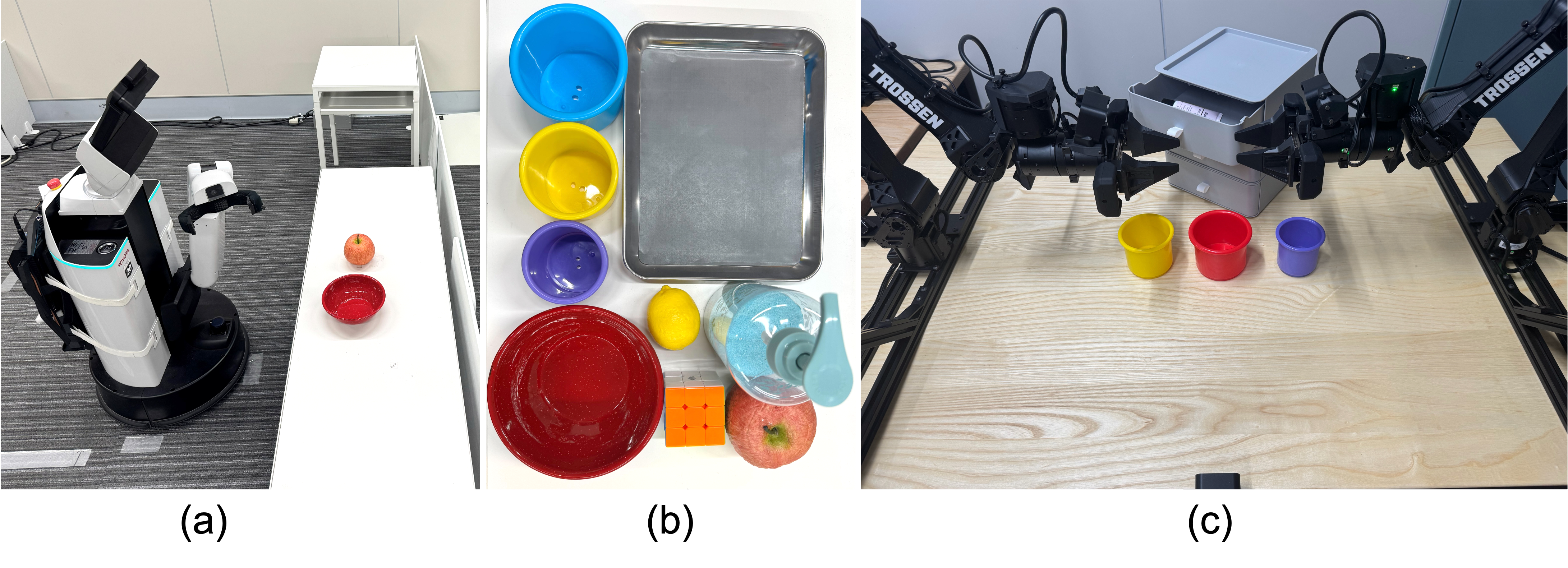}
  \caption{
        (a) Robot platform and experimental environment used in the mobile manipulation experiments. We used the Human Support Robot (HSR) as the robot platform. 
        (b) Everyday objects used in the experiments. We used objects with diverse appearances, shapes, and sizes to emulate common household manipulation scenarios. (c) {ALOHA robot platform and experimental setup used in the tabletop manipulation experiments.} 
    }
    \label{fig:setting}
\end{figure}
\subsubsection{{Mobile Manipulation Experiments.}}
Fig. \ref{fig:setting} illustrates the experimental setup for the real-world experiments. 
We constructed an environment with furniture and objects to emulate domestic manipulation scenarios, as shown in Fig.~\ref{fig:setting}(a).
Furthermore, we used an additional camera to capture a bird's-eye view.

In the mobile manipulation experiments, we used the Human Support Robot developed by Toyota Motor Corporation.
The robot features a total of 11 degrees of freedom (DoF), comprising 3 DoF in the mobile base, 6 DoF in the manipulator, and 2 DoF in the head. Furthermore, it is equipped with cameras mounted on both its head and end-effector.
We used a variety of everyday objects with different appearances, shapes, and sizes, as shown in Fig. \ref{fig:setting}(b).
In the mobile manipulation experiments, we evaluated AnolisVLA across five representative mobile manipulation tasks to demonstrate its versatility.
The definitions of these tasks are as follows: `Mobile move': Move an object laterally. `Mobile pick': Grasp and lift an object. `Mobile open': Open a drawer. `Mobile close': Close a drawer. `Mobile push': Depress a bottle pump.


{
This setting is more challenging than tabletop manipulation because the camera viewpoint changes as the robot’s mobile base moves during manipulation.
Consequently, it poses a significant challenge for representative VLA models (e.g., $\pi_{0.5}$), which are largely pretrained on data collected in tabletop manipulation setups and therefore assume that the viewpoint remains stable.
}


We used a leader-follower system \cite{takanami2025airoa} for data collection. 
A human operator manipulated a leader device to teleoperate the robot and collect trajectories. 
We collected 50 episodes for each task while randomly varying object placement and the robot's initial pose, which resulted in a dataset totaling 250 episodes. 
We recorded each episode at a sampling rate of 10 Hz and took images from three viewpoints (external, head-mounted, and wrist-mounted cameras), along with all joint angles at each timestep.
We set $d_a$ to 11.

\subsubsection{{Tabletop Manipulation Experiments.}}
{Fig. \ref{fig:setting}(c) illustrates the experimental setup for the tabletop manipulation experiments. We used an ALOHA-style bimanual manipulation platform~\cite{ACT}, which consists of two follower arms for task execution and two leader arms for teleoperation. This setting complements the mobile manipulation experiments by evaluating whether AnolisVLA can generate coordinated trajectories for two arms in tabletop manipulation.
We evaluated AnolisVLA on two bimanual manipulation tasks. The definitions of these tasks are as follows: `Stack cup': Insert the two side cups into the center cup in the specified order. `Put in drawer': Place an object into a drawer and then close the drawer. We used the leader-follower teleoperation interface of ALOHA for data collection.  We collected 100 episodes for each task. Each episode was recorded at a sampling rate of 30 Hz using multi-view RGB images, along with the joint angles of both arms. We set $d_a$ to 14.
}

\subsection{Quantitative Results}

\begin{table}[t]
    \centering
    \setlength{\tabcolsep}{2pt}
    \caption{Quantitative comparison between AnolisVLA and the baseline methods in mobile manipulation experiments.
    The best results are highlighted in \textbf{bold}, and the second-best results are \underline{underlined}.}
    \label{tab:realworld_results}
    \scalebox{0.95}{
    \begin{tabular}{l ccccc c}
        \toprule
        Method & \multicolumn{6}{c}{Success rate [\%]} \\
        \cmidrule(lr){2-7}
        & Mobile move & Mobile pick & Mobile open & Mobile close & Mobile push & Avg. \\
        \midrule
        $\pi_{0.5}$ \cite{pi05} & 15 & 20 & 30 & \textbf{100} & 45 & 42 \\
        \midrule
        VLA-Adapter \cite{VLA-Adapter} & \underline{35} & 20 & 0 & \textbf{100} & 45 & 40 \\
        TinyVLA \cite{TinyVLA} & 5 & 15 & 0 & \underline{90} & 35 & 29 \\
        SmolVLA \cite{SmolVLA} & 30 & \underline{25} & \underline{55} & \textbf{100} & \underline{50} & \underline{52} \\
        \rowcolor{green!10}
        \textbf{AnolisVLA} & \textbf{45} & \textbf{40} & \textbf{75} & \textbf{100} & \textbf{55} & \textbf{63} \\
        \bottomrule
    \end{tabular}
    }
\end{table}
\begin{table}[t]
    \centering
    \setlength{\tabcolsep}{4pt}
    \caption{Quantitative comparison between AnolisVLA and the baseline methods on the ALOHA robot, including task success rates, inference time, and inference-time VRAM consumption. The `VRAM' column reports the inference-time GPU memory consumption. The best results are highlighted in \textbf{bold}, and the second-best results are \underline{underlined}.}
    \label{tab:aloha_results}
    \begin{tabular}{l cc c c c}
        \toprule
        Method & \multicolumn{3}{c}{Success rate [\%]} & \shortstack{Inference time  $\downarrow$} & \shortstack{VRAM  $\downarrow$} \\
        \cmidrule(lr){2-4}
        & Stack cup & Put in drawer & Avg. & [ms/chunk] & [GB]\\
        \midrule
        $\pi_{0.5}$ \cite{pi05} & \textbf{70} & \textbf{80} & \textbf{75} & 578 & 17.05 \\
        \midrule
        VLA-Adapter \cite{VLA-Adapter} & 50 & 30 & 40 & \textbf{101} & 4.65 \\
        TinyVLA \cite{TinyVLA} & 30 & 40 & 35 & 1290 & 2.38 \\
        SmolVLA \cite{SmolVLA} & 40 & \underline{70} & 55 & 309 & \textbf{1.56} \\
        \rowcolor{green!10}
        \textbf{AnolisVLA} & \underline{60} & \textbf{80} & \underline{70} & \underline{216} & \underline{1.64} \\
        \bottomrule
    \end{tabular}
\end{table}

Table~\ref{tab:realworld_results} shows a quantitative comparison between AnolisVLA and the baseline methods in the mobile manipulation experiments.
For each task, we conducted 20 trials. The resulting success rates are presented in the table.

AnolisVLA achieved an average success rate of 63\%, whereas the values of VLA-Adapter, TinyVLA, and SmolVLA were 40\%, 29\%, and 52\%, respectively.
Consequently, AnolisVLA outperformed SmolVLA, the strongest among the small-scale VLAs, by 11 points for the average success rate.
Furthermore, although SmolVLA performed moderately well overall, it showed lower performance in spatially constrained, contact-rich tasks, achieving a success rate of 55\% on the ‘Open’ task. 
By contrast, AnolisVLA achieved a success rate of 75\% on this task. 
This result suggests that AnolisVLA was effective on this contact-rich drawer-opening task.

The relatively low success rate of $\pi_{0.5}$ in the mobile manipulation experiments can be mainly attributed to the difficulty of this setting.
Unlike standard tabletop manipulation, the robot executes each task from different base positions, which changes the camera viewpoints and the spatial relationship between the robot and the target objects. 
Thus, the policy must handle viewpoint-dependent object recognition and whole-body motion in addition to object manipulation.
This introduces a substantial domain gap from fixed-view tabletop demonstrations, on which large-scale VLAs such as $\pi_{0.5}$ are mainly trained. 
As a result, even a strong general-purpose VLA can exhibit degraded performance in spatially constrained and contact-rich mobile manipulation tasks.
Importantly, this result should be interpreted in light of the evaluation setting rather than as a general limitation of $\pi_{0.5}$. As shown later in Table~\ref{tab:aloha_results}, $\pi_{0.5}$ achieved the highest average success rate in the tabletop bimanual setting. 

Table~\ref{tab:aloha_results} shows the quantitative comparison on the ALOHA robot, including the task success rate, inference time, and inference-time VRAM consumption. For each task, we conducted 10 trials. AnolisVLA achieved success rates of 60\% and 80\% on the `Stack cup' and `Put in drawer' tasks, respectively, resulting in an average success rate of 70\%. Among the small-scale VLAs, AnolisVLA achieved the highest average success rate, outperforming SmolVLA by 15 points. Across all methods, however, $\pi_{0.5}$ achieved an average success rate of 75\%. These results indicate that AnolisVLA is effective in tabletop  manipulation while providing a favorable performance-efficiency tradeoff among lightweight VLAs.

Regarding inference time, AnolisVLA achieved 216~ms/chunk.
SmolVLA and $\pi_{0.5}$ took 309 and 578~ms/chunk, respectively.
At a comparable parameter scale, AnolisVLA was 1.43 times faster than SmolVLA.
In a comparison between models of different parameter scales, AnolisVLA was approximately three times as fast as $\pi_{0.5}$.
VLA-Adapter achieved the lowest latency at 101~ms/chunk in our setup.
However, it yielded substantially lower success rates, which indicates that AnolisVLA provided a better balance between the success rate and inference time under limited compute budgets.

{Regarding inference-time GPU memory usage, AnolisVLA required 1.64 GB of VRAM.
This was substantially smaller than $\pi_{0.5}$, which required 17.05 GB, corresponding
to a 10.4-fold reduction in VRAM usage. Although SmolVLA used
slightly less VRAM than AnolisVLA , AnolisVLA achieved higher task success while
maintaining comparable memory consumption. These results indicate that AnolisVLA 
provides a favorable tradeoff among task success, inference latency, and GPU memory
usage.}


\begin{figure}[tb]
  \centering
  \includegraphics[  width=0.9\textwidth,
  keepaspectratio]{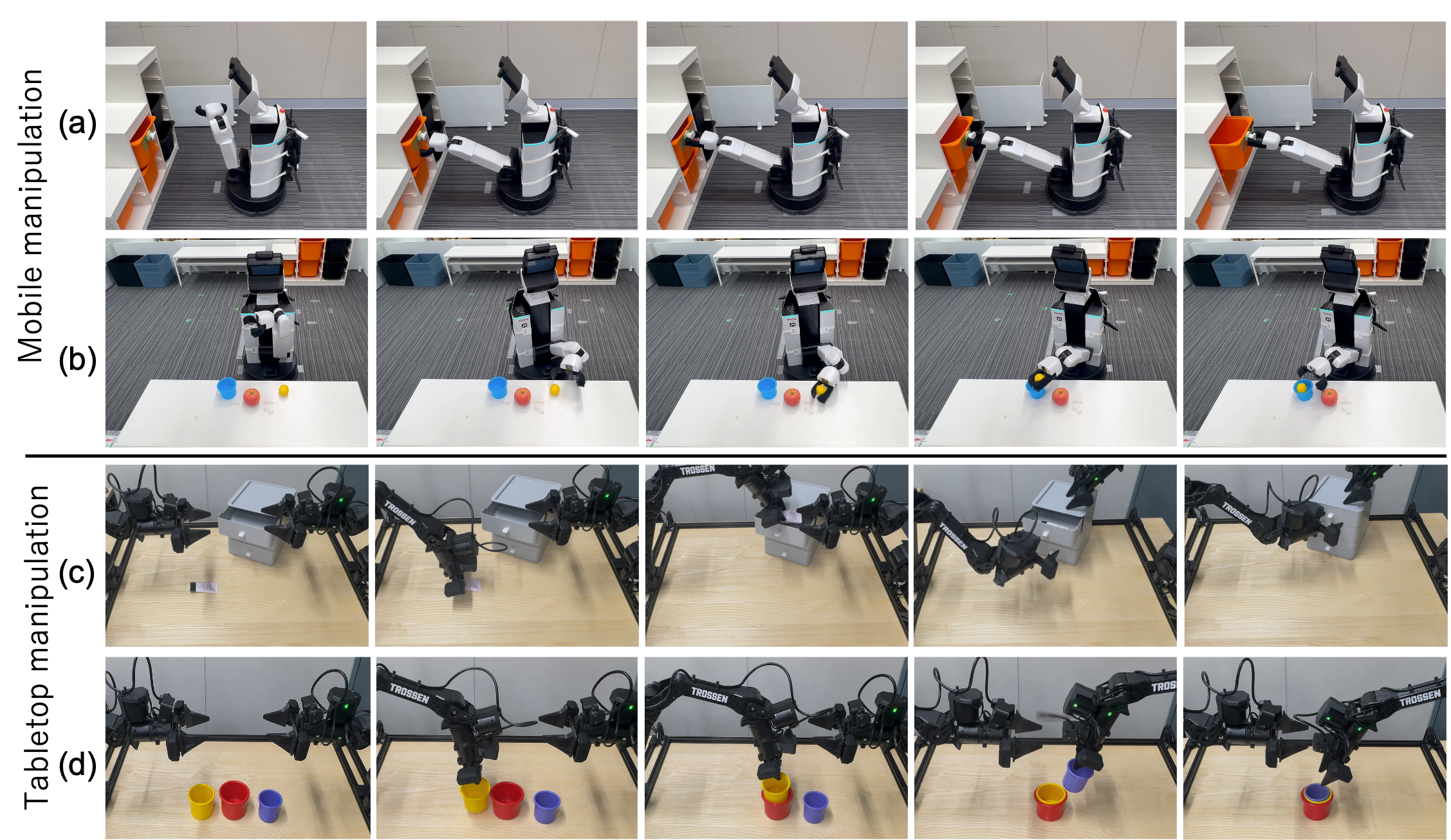}
  \caption{
        Successful episodes of AnolisVLA in the real-world experiments. Instructions were as follows: (a) ``Open the drawer.'' (b) ``Put the lemon into the cup.'' {(c) ``Put the object in the drawer.'' (d) ``Stack the cups.''}
    }
    \label{fig:qualitative_real-world}
\end{figure}

\subsection{Qualitative Results}
Fig.~\ref{fig:qualitative_real-world} shows selected successful manipulation episodes with AnolisVLA in the real-world experiments. Each sequence of images illustrates the observations in chronological order from left to right. The upper two rows show examples from the mobile manipulation experiments, whereas the lower two rows show examples from the tabletop manipulation experiments.

In episode (a), given the instruction ``Open the drawer,'' the mobile manipulator grasped the handle and pulled it forward to open the drawer. In episode (b), given the instruction ``Put the lemon into the cup,'' the robot lifted the lemon and placed it into the target cup. These examples show that AnolisVLA can generate mobile manipulation trajectories from language instructions under changing camera viewpoints.
In episode (c), given the instruction ``Put the object in the drawer,'' the ALOHA robot placed the target object into the drawer and then closed it. In episode (d), given the instruction ``Stack the cups,'' the robot inserted the two side cups into the center cup in the specified order. These examples show that AnolisVLA can also generate coordinated bimanual trajectories for tabletop manipulation tasks.

\section{Limitation and Conclusion}
Our error analysis identified incorrect target localization, inaccurate grasp-point prediction, and incomplete motion execution as the main failure modes.
Furthermore, our deterministic single-pass regression cannot explicitly represent multiple valid action modes for the same input.

In this study, we addressed language-guided robotic manipulation, where a robot performs object manipulation based on visual observations and language instructions.
Our contributions are fourfold: 
(i) we proposed AnolisVLA, a lightweight and high-speed VLA for resource-constrained environments; 
(ii) we instantiated it with a Mamba backbone for efficient VLA-based trajectory generation; 
(iii) we introduced two-stage training with temporal-difference supervision; and 
(iv) we evaluated it in simulation and on two real-world robot platforms, demonstrating its effectiveness in mobile and bimanual tabletop manipulation.


\paragraph{{\bf Acknowledgments.}}
This work was partially supported by JSPS KAKENHI Grant Number 23K28168,  JST CREST, and JSPS Fellows Grant Number JP23KJ1917.

\bibliography{reference}  

@String(CVPR  = {IEEE Conf. Comput. Vis. Pattern Recog.})

@String(NeurIPS = {Adv. Neural Inform. Process. Syst.})

@String(ICLR  = {Int. Conf. Learn. Represent.})

@String(AAAI  = {AAAI})

@String(ICASSP=	{ICASSP})

@String(CVPR  = {CVPR})

@String(NeurIPS = {NeurIPS})

@String(ICLR  = {ICLR})

@article{siglip_2,
  title={{SigLIP 2: Multilingual Vision-Language Encoders with Improved Semantic Understanding, Localization, and Dense Features}},
  author={Tschannen, Michael and Gritsenko, Alexey and others},
  journal={arXiv preprint arXiv:2502.14786},
  year={2025}
}

@inproceedings{Brohan2023rt2,
  title={{RT-2: Vision-Language-Action Models Transfer Web Knowledge to Robotic Control}},
  author={Zitkovich, Brianna and Yu, Tianhe and Xu, Sichun and others},
  booktitle={CoRL},
  year={2023},
}

@inproceedings{Kim2024openvla,
  title={{OpenVLA: An Open-Source Vision-Language-Action Model}},
  author={Kim, Moo and Pertsch, Karl and Karamcheti, Siddharth and others},
  booktitle={CoRL},
  year={2024}
}

@inproceedings{pi0,
  title={$\pi_0$: {A Vision-Language-Action Flow Model for General Robot Control}},
  author={Black, Kevin and Brown, Noah and Driess, Danny and others},
  booktitle={RSS},
  year={2025}
}

@inproceedings{pi05,
  title={$\pi_{0.5}$: {A Vision-Language-Action Model with Open-World Generalization}},
  author={Black, Kevin and Brown, Noah and Darpinian, James and others},
  booktitle={CoRL},
  year={2025}
}

@inproceedings{UniVLA,
  title={{UniVLA: Learning to Act Anywhere with Task-centric Latent Actions}},
  author={Bu, Qingwen and Yang, Yanting and Cai, Jisong and Gao, Shenyuan and Ren, Guanghui and Yao, Maoqing and Luo, Ping and Li, Hongyang},
  booktitle={RSS},
  year={2025}
}

@inproceedings{X-VLA,
  title   = {{X-VLA: Soft-Prompted Transformer as Scalable Cross-Embodiment Vision-Language-Action Model}},
  author  = {Zheng, Jinliang and Li, Jianxiong and Wang, Zhihao and Liu, Dongxiu and Kang, Xirui and Feng, Yuchun and Zheng, Yinan and Zou, Jiayin and Chen, Yilun and others},
  booktitle = {ICLR},
  year    = {2026}
}

@inproceedings{octo,
    title={{Octo: An Open-Source Generalist Robot Policy}},
    author = {Dibya Ghosh and Homer Walke and Karl Pertsch and Kevin Black and Oier Mees and Sudeep Dasari and Joey Hejna and Charles Xu and others},
    booktitle = {RSS},
    year = {2024},
}

@inproceedings{OXE,
  title={{Open X-Embodiment: Robotic Learning Datasets and RT-X Models: Open X-Embodiment Collaboration}},
  author={O'Neill, Abby and others},
  booktitle={ICRA},
  year={2024},
  pages={6892--6903},
}

@inproceedings{VoxPoser,
  title={{VoxPoser: Composable 3D Value Maps for Robotic Manipulation with Language Models}},
  author={Huang, Wenlong and Wang, Chen and Zhang, Ruohan and others},
  booktitle={CoRL},
  year={2023},
  pages={540--562},
}

@inproceedings{RT-1,
  title={{RT-1: Robotics Transformer for Real-World Control at Scale}},
  author={Brohan, Anthony and Brown, Noah and Carbajal, Justice and Chebotar, Yevgen and others},
  booktitle={RSS},
  year={2023}
}

@article{TinyVLA,
  title={{TinyVLA: Towards Fast, Data-Efficient Vision-Language-Action Models for Robotic Manipulation}},
  author={Wen, Junjie and Zhu, Yichen and others},
  journal={IEEE RA-L},
  year={2025},
  publisher={IEEE}
}

@inproceedings{BridgeV2,
  title={{BridgeData V2: A Dataset for Robot Learning at Scale}},
  author={Walke, Homer and Black, Kevin and Lee, Abraham and Kim, Moo Jin and Du, Max and Zheng, Chongyi and Zhao, Tony and others},
  booktitle={CoRL},
  year={2023}
}

@inproceedings{Yu2019MetaWorld,
  title={{Meta-World: A Benchmark and Evaluation for Multi-Task and Meta Reinforcement Learning}},
  author={Yu, Tianhe and Quillen, Deirdre and He, Zhanpeng and others},
  booktitle={CoRL},
  year={2019}
}

@inproceedings{VLA-Adapter,
  title     = {{VLA-Adapter}: An Effective Paradigm for Tiny-Scale Vision-Language-Action Model},
  author    = {Wang, Yihao and Ding, Pengxiang and Li, Lingxiao and Cui, Can and Ge, Zirui and Tong, Xinyang and Song, Wenxuan and Zhao, Han and Zhao, Wei and Hou, Pengxu and others},
  booktitle = {AAAI},
  volume    = {40},
  number    = {22},
  pages     = {18638--18646},
  year      = {2026},
}

@inproceedings{RoboMamba,
  title={{RoboMamba: Efficient Vision-Language-Action Model for Robotic Reasoning and Manipulation}},
  author={Liu, Jiaming and Liu, Mengzhen and Wang, Zhenyu and An, Pengju and others},
  booktitle={{NeurIPS}},
  year={2024}
}

@article{SmolVLA,
  title={{SmolVLA: A Vision-Language-Action Model for Affordable and Efficient Robotics}},
  author={Shukor, Mustafa and Aubakirova, Dana and Capuano, Francesco and Kooijmans, Pepijn and Palma, Steven and others},
  journal={arXiv preprint arXiv:2506.01844},
  year={2025}
}

@inproceedings{SpatialVLA,
  title={{SpatialVLA: Exploring Spatial Representations for Visual-Language-Action Model}},
  author={Qu, Delin and Song, Haoming and Chen, Qizhi and Yao, Yuanqi and Ye, Xinyi and Ding, Yan and Wang, Zhigang and Gu, JiaYuan and Zhao, Bin and Wang, Dong and others},
  booktitle={RSS},
  year={2025}
}

@inproceedings{CoT-VLA,
  title={{CoT-VLA: Visual Chain-of-Thought Reasoning for Vision-Language-Action Models}},
  author={Zhao, Qingqing and Lu, Yao and Kim, Moo and Fu, Zipeng and Zhang, Zhuoyang and Wu, Yecheng and Li, Zhaoshuo and Ma, Qianli and Han, Song and Finn, Chelsea and others},
  booktitle={CVPR},
  year={2025},
  pages={1702--1713},
}

@inproceedings{TraceVLA,
  title={{TraceVLA: Visual Trace Prompting Enhances Spatial-Temporal Awareness for Generalist Robotic Policies}},
  author={Ruijie Zheng and Yongyuan Liang and others},
  booktitle={ICLR},
  year={2025},
}

@article{Ma2024,
  title={{A Survey on Vision-Language-Action Models for Embodied AI}},
  author={Ma, Yueen and Song, Zixing and Zhuang, Yuzheng and Hao, Jianye and King, Irwin},
  journal={arXiv preprint arXiv:2405.14093},
  year={2024}
}

@inproceedings{Kawaharazuka2023,
  title={{Robotic Applications of Pre-Trained Vision-Language Models to Various Recognition Behaviors}},
  author={Kawaharazuka, Kento and others},
  booktitle={Humanoids},
  year={2023},
  pages={458--465},
}

@inproceedings{Hori,
  title={{Interactive Robot Action Replanning using Multimodal LLM Trained from Human Demonstration Videos}},
  author={Hori, Chiori and Kambara, Motonari and Sugiura, Komei and Ota, Kei and Khurana, Sameer and Jain, Siddarth and others},
  booktitle={ICASSP},
  pages={1--5},
  year={2025},
}

@article{VLMPC,
  title={{VLMPC: Vision-Language Model Predictive Control for Robotic Manipulation}},
  author={Zhao, Wentao and Chen, Jiaming and Meng, Ziyu and others},
  journal={arXiv preprint arXiv:2407.09829},
  year={2024}
}

@inproceedings{Zhao2025cobra,
      title={{Cobra: Extending Mamba to Multi-Modal Large Language Model for Efficient Inference}}, 
      author={Han Zhao and Min Zhang and Wei Zhao and Pengxiang Ding and others},
      booktitle={AAAI},
      year={2025},
}

@article{Qiao2024vlmamba,
  author = {Qiao, Yanyuan and Yu, Zheng and Guo, Longteng and Chen, Sihan and others},
  title  = {{VL-Mamba: Exploring State Space Models for Multimodal Learning}},
  journal = {arXiv preprint arXiv:2403.13600},
  year   = {2024}
}

@inproceedings{Xing2025emma,
  title={{EMMA: Empowering Multi-modal Mamba with Structural and Hierarchical Alignment}},
  author={Xing, Yifei and Lan, Xiangyuan and Wang, Ruiping and Jiang, Dongmei and others},
  booktitle={ICLR},
  year={2025}
}

@inproceedings{gu2021efficiently,
  title={{Efficiently Modeling Long Sequences with Structured State Spaces}},
  author={Gu, Albert and Goel, Karan and R{\'e}, Christopher},
  booktitle={ICLR},
  year={2022}
}

@inproceedings{gu2022parameterization,
 author = {Gu, Albert and Goel, Karan and Gupta, Ankit and R\'{e}, Christopher},
 booktitle = {NeurIPS},
 pages = {35971--35983},
 title = {{On the Parameterization and Initialization of Diagonal State Space Models}},
 volume = {35},
 year = {2022}
}

@inproceedings{gu2024mamba,
  title={{Mamba: Linear-Time Sequence Modeling with Selective State Spaces}},
  author={Gu, Albert and Dao, Tri},
  booktitle={CoLM},
  year={2024}
}

@article{takanami2025airoa,
  title={{AIRoA MoMa Dataset: A Large-Scale Hierarchical Dataset for Mobile Manipulation}},
  author={Takanami, Ryosuke and others},
  journal={arXiv preprint arXiv:2509.25032},
  year={2025}
}

@inproceedings{openvla-oft,
  title     = {{Fine-Tuning Vision-Language-Action Models: Optimizing Speed and Success}},
  author    = {Kim, Moo Jin and Finn, Chelsea and Liang, Percy},
  booktitle = {RSS},
  year      = {2025}
}

@inproceedings{FlowRAM,
  title     = {{FlowRAM: Grounding Flow Matching Policy with Region-Aware Mamba Framework for Robotic Manipulation}},
  author    = {Wang, Sen and Wang, Le and Zhou, Sanping and Tian, Jingyi and Li, Jiayi and Sun, Haowen and others},
  booktitle = {CVPR},
  year      = {2025},
  pages     = {12176--12186}
}

@inproceedings{LUMOS,
  title     = {{LUMOS: Language-Conditioned Imitation Learning with World Models}},
  author    = {Nematollahi, Iman and DeMoss, Branton and Chandra, Akshay and others},
  booktitle = {ICRA},
  year      = {2025},
  pages     = {8219--8225}
}

@inproceedings{MaIL,
  title     = {{MaIL: Improving Imitation Learning with Selective State Space Models}},
  author    = {Jia, Xiaogang and Wang, Qian and Donat, Atalay and Xing, Bowen and Li, Ge and Zhou, Hongyi and Celik, Onur and Blessing, Denis and Lioutikov, Rudolf and Neumann, Gerhard},
  booktitle = {CoRL},
  year      = {2025},
  pages     = {3888--3907}
}

@inproceedings{RoboSSM,
  title     = {{RoboSSM: Scalable In-context Imitation Learning via State-Space Models}},
  author    = {Yoo, Youngju and Hu, Jiaheng and Zhu, Yifeng and Liu, Bo and Liu, Qiang and others},
  booktitle = {IROS},
  year      = {2026},
}

@inproceedings{cao2025mamba,
  title={{Mamba Policy: Towards Efficient 3D Diffusion Policy with Hybrid Selective State Models}},
  author={Cao, Jiahang and Zhang, Qiang and Sun, Jingkai and others},
  booktitle={IROS},
  pages={11359--11366},
  year={2025}
}

@inproceedings{oh2024dispo,
  title={{DiSPo: Diffusion-SSM based Policy Learning for Coarse-to-Fine Action Discretization}},
  author={Oh, Nayoung and Jang, Jaehyeong and Jung, Moonkyeong and Park, Daehyung},
  booktitle={ICRA},
  year={2026}
}

@article{tsuji2025mamba,
  title={{Mamba as a Motion Encoder for Robotic Imitation Learning}},
  author={Tsuji, Toshiaki},
  journal={IEEE Access},
  volume={13},
  pages={69941--69949},
  year={2025}
}

@inproceedings{ACT,
  title     = {{Learning Fine-Grained Bimanual Manipulation with Low-Cost Hardware}},
  author    = {Zhao, Tony and others},
  booktitle = {Proceedings of Robotics: Science and Systems},
  year      = {2023}
}





\clearpage  

%
%
\bibliographystyle{splncs04}
\end{document}